\documentclass{article}

\usepackage{arxiv}

\usepackage[utf8]{inputenc} 
\usepackage[T1]{fontenc}    
\usepackage{hyperref}       
\usepackage{url}            
\usepackage{booktabs}       
\usepackage{amsfonts}       
\usepackage{nicefrac}       
\usepackage{microtype}      
\usepackage{lipsum}		
\usepackage{graphicx}
\usepackage{doi}
\usepackage{comment}
\usepackage{amsfonts}
\usepackage{amssymb}
\usepackage{amsmath}
 \usepackage[numbers]{natbib}
\title{Time-adaptive Video Frame Interpolation based on Residual Diffusion}

\author{ Victor Fonte Chavez \\
	Department of Computer Science\\
	Centro de Investigación en Matemáticas\\
	Guanajuato, México \\
	\texttt{victor.fonte@cimat.mx} \\
	\And
	\href{https://orcid.org/0000-0002-3323-0510}{\includegraphics[scale=0.06]{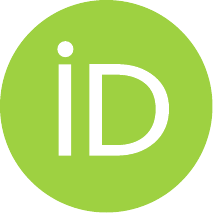}\hspace{1mm}Claudia Esteves} \\
	Department of Mathematics\\
	university of Guanajuato\\
	Guanajuato, México \\
	\texttt{cesteves@cimat.x} \\
	\And
	\href{https://orcid.org/0000-0002-3323-0510}{\includegraphics[scale=0.06]{images/orcid.pdf}\hspace{1mm}Jean-Bernard Hayet} \\
	Department of Computer Science\\
	Centro de Investigación en Matemáticas\\
	Guanajuato, México \\
	\texttt{jbhayet@cimat.x} \\
}

\renewcommand{\shorttitle}{Time-adaptive Video Frame Interpolation based on Residual Diffusion}

\hypersetup{
pdftitle={A template for the arxiv style},
pdfsubject={q-bio.NC, q-bio.QM},
pdfauthor={David S.~Hippocampus, Elias D.~Striatum},
pdfkeywords={First keyword, Second keyword, More},
}

\usepackage{subcaption}
\usepackage{tabularx}
\newcolumntype{Y}{>{\centering\arraybackslash}X}

\DeclareMathOperator{\lpips}{lpips}

\begin{document}

\maketitle

\begin{abstract}
In this work, we propose a new diffusion-based method for video frame interpolation (VFI), in the context of traditional hand-made animation. We introduce three main contributions: The first is that we explicitly handle the interpolation time in our model, which we also re-estimate during the training process, to cope with the particularly large variations observed in the animation domain, compared to natural videos; The second is that we adapt and generalize a diffusion scheme called ResShift recently proposed in the super-resolution community to VFI, which allows us to perform a very low number of diffusion steps (in the order of $10$) to produce our estimates; The third is that we leverage the stochastic nature of the diffusion process to provide a pixel-wise estimate of the uncertainty on the interpolated frame, which could be useful to anticipate where the model may be wrong. We provide extensive comparisons with respect to state-of-the-art models and show that our model outperforms these models on animation videos. Our code is available at \href{https://github.com/VicFonch/Multi-Input-Resshift-Diffusion-VFI}{https://github.com/VicFonch/Multi-Input-Resshift-Diffusion-VFI}
\end{abstract}

\keywords{Video frame interpolation, Diffusion models, Deep learning}


\begin{figure}[h]
    \centering
    \begin{tabular}{ccccc}
        \includegraphics[height=2cm]{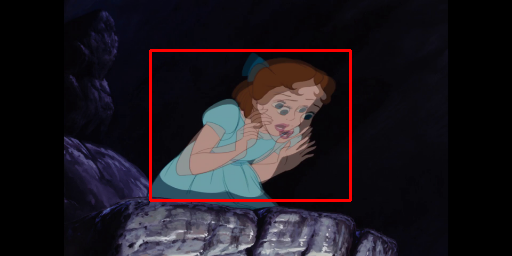} &
        \includegraphics[height=2cm]{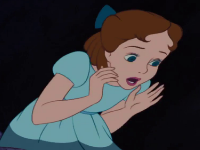} &
        \includegraphics[height=2cm]{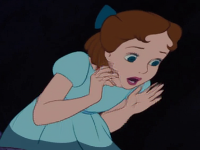} &
        \includegraphics[height=2cm]{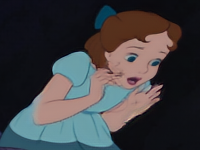} &
        \includegraphics[height=2cm]{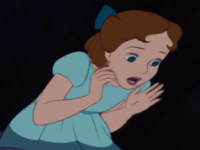} \\
        $\mathbf I_0 + \mathbf I_1$ & GT & Eisei & SoftSplat & Ours \\
    \end{tabular}
    \caption{Zoomed-in qualitative comparison on a challenging interpolation scenario. From left to right: initial frame $\mathbf I_0$, final frame $\mathbf I_1$, ground truth (cropped), SoftSplat prediction, and our result. Note the quality of the reconstructed fingers for the different methods.}
    \label{fig:zoom-comparison}
\end{figure}

\section{Introduction}

Traditional 2D animation requires a significant manual effort, as animators must create approximately twelve individual illustrations for every second of movement. This workflow demands large production teams, each with specialized training to generate and colorize a massive number of frames needed to produce an animated short film. With the growing global demand for traditional animation, studios face increasing pressure to produce high-quality content efficiently. In this context, Video Frame Interpolation (VFI) emerges as a key possible tool, with applications ranging from motion smoothing to frame rate enhancement. Traditional VFI methods based on optical flow and motion estimation struggle with occlusions, lighting variations, and complex motion, leading to artifacts such as blurring and ghosting. Recent advances in deep learning, particularly diffusion models, offer a promising alternative by effectively approximating high-dimensional data distributions. This work proposes a computationally efficient diffusion-based VFI model that integrates with traditional methods. Additionally, we propose to estimate and encode systematically the temporal position of the intermediate frame, mitigating training biases and inaccuracies and enabling flexible interpolation at arbitrary positions. Our approach aims to establish a scalable and adaptable diffusion framework for frame interpolation, addressing key limitations of existing techniques and paving the way for real-time applications.

\section{Video Frame Interpolation: Overview of previous work}

For a few years now, deep learning-based methods have been forming the bulk of the state of the art in VFI, with most works using convolutional, encoder-decoder architectures to perform interpolation. We describe some of them hereafter.

In~\cite{Voxel-Flow_original}, to synthesize the video frames, Deep Voxel Flow (DVF) interpolates the pixel values from the frames that are close to the one to interpolate, by applying a voxel flow layer across space and time in the input video. Trilinear interpolation across the input video volume generates the final pixel value. The deep bidirectional predictive network (BiPN)~\cite{Bidirectional-Net_original} is a convolutional encoder-decoder network trained to regress the missing intermediate frames from two opposite directions, with a bi-directional encoder-decoder that simultaneously predicts the future-forward from the starting frame and predicts the past-backward from the ending frame. Multiple missing frames can be predicted by the decoder after taking the feature representations as input.

PhaseNet~\cite{Phasenet_original} estimates the phase decomposition of the intermediate frame. It is designed as a decoder-only network, increasing its resolution level by level. The input is the response from the steerable pyramid decomposition of the two input frames, consisting of the phase and amplitude values for each pixel at each level. Each resolution level has a PhaseNet block that takes the decomposition values from the input images as its input, altogether with the resized feature maps and the resized predicted values from the previous level and it outputs the decomposition values of the intermediate image, from which the intermediate image is reconstructed. 

In~\cite{Super-slomo_original}, the authors propose a solution for variable-length multi-frame video interpolation, with motion interception and occlusion jointly modeled. Bidirectional optical flows between input images are computed using a U-Net and are linearly combined to approximate the intermediate optical flows. The approximated optical flows are refined using another U-Net that also predicts soft visibility maps. The two input images are warped and linearly joined to form intermediate frames. In~\cite{Depth-Aware-VFI_original}, the Depth-Aware video frame Interpolation (DAIN) model detects occlusion through depth information. A depth-aware flow projection layer synthesizes flows that sample closer objects more often than those far away. The output frame is generated by warping the input frames, depth maps, and contextual features based on the optical flow and the local interpolation kernels.  
 
In~\cite{VFI-Adaptive-Separable-Convolution_original,VFI-Adaptive-Convolution_original}, the authors introduce a spatially-adaptive separable convolution technique to interpolate the intermediate frames. Pixelwise, it estimates a pair of 2D convolution kernels (four 1D kernels) to convolve the two video frames and compute the color of the output pixel.
The pixel-dependent kernels capture both motion and re-sampling information required for interpolation. This idea is known as \emph{soft-splatting} and has been at the core of many VFI methods. In~\cite{DSepConv_original}, the authors improve the soft-splatting idea with Adaptive Deformable Separable Convolution to adaptively estimate kernels, offsets and masks so that the network obtains information with fewer but more relevant pixels than in~\cite{VFI-Adaptive-Separable-Convolution_original}. The learnable offsets make that pixels outside the local neighborhood can be reached, allowing to better handle large motion with smaller convolution kernels. 

As in many areas, generative modeling has been used in VFI, which can be cast as a conditional image generation problem. FIGAN~\cite{GAN-VFI_original} is a multi-scale generative adversarial network for frame interpolation, where the predicted flow and synthesized frame are constructed in a coarse-to-fine fashion. The network is jointly supervised at different levels with an adversarial and two content losses. A refinement module jointly processes the synthesized image with the original input frames that produced it. LDMVFI~\cite{LDMVFI_original} is a latent diffusion model adapted from the latent diffusion models~\cite{LDM_original}. It includes an autoencoding model that projects images into a latent space, and a denoising U-Net that performs reverse diffusion in that latent space. The encoder to the latent space is a VQ-VAE that differs from the original VQ-GAN~\cite{VQGAN_original} in its use of MaxViT-based cross attention~\cite{MaxViT_original} for the feature extraction and of adaptive deformable convolution like~\cite{DSepConv_original}. 

Given the great performance obtained by diffusion models in image generation, in this work we consider adapting this type of models to the VFI problem. Other works have leveraged diffusion models in VFI such as~\cite{LDMVFI_original} or~\cite{Google_VIDIM_original}, but they share a characteristic coming from the original diffusion model: Their slowness due to the massive number of forward evaluations from samplers like DDPM~\cite{DDPM_original} and the gigantic number of parameters needed to get realistic results from the denoiser models. Therefore, in this work, our focus is to propose a diffusion model that is efficient in size and time and is, at the same time, capable of reaching state-of-the-art (SOTA) models. Because of its remarkable improvements in the sampling process, we base our proposal on ResShift, a diffusion scheme proposed originally in the image super-resolution problem~\cite{Resshift_original}.

\section{A time-adaptive diffusion model for frame interpolation}

\begin{figure}[t]
    \centering
    \includegraphics[width=0.48\linewidth]{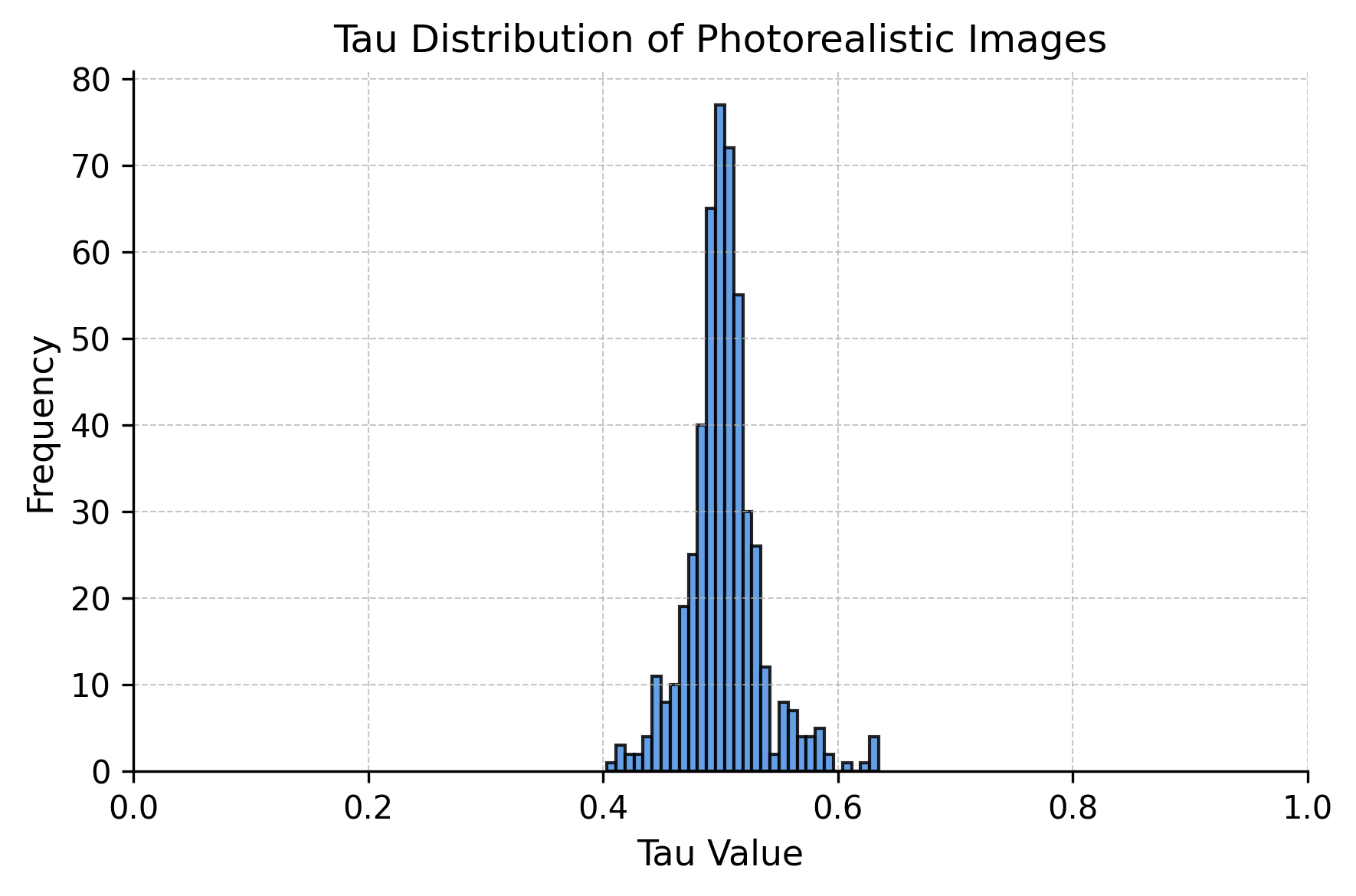}
    \includegraphics[width=0.48\linewidth]{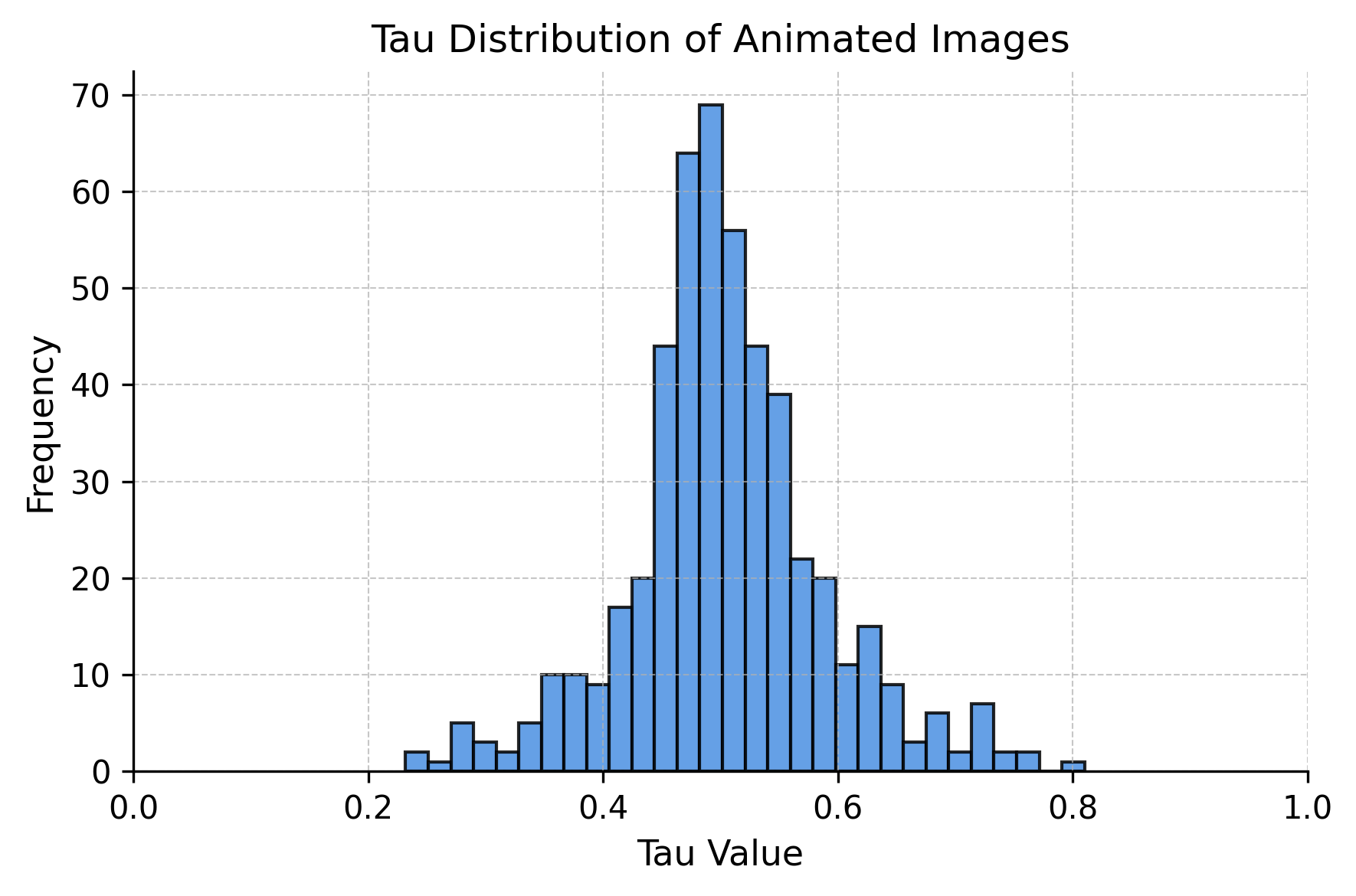}
    \caption{Distribution of $\tau$ for photorealistic images (left) and animated images (right). The histograms show a concentration of values around $0.5$ in both categories, with a much higher dispersion in the animated images.}
    \label{fig:distro-tau}
\end{figure}

Given a pair of consecutive frames $\mathbf{I}_0$ and $\mathbf{I}_1$ from an original video, the goal of VFI is to estimate (interpolate) intermediate frames $\mathbf{I}_{\tau}$ for arbitrary values of $\tau \in (0,1)$. Let us emphasize that this problem takes a particular flavor in hand-made animation. Assuming that the motion is uniform along time, within a natural video, one has that the intermediate frame $\tau=\frac{1}{2}$ should correspond to half of the motion between $\mathbf{I}_0$ and $\mathbf{I}_1$. In the case of animation, this correspondence is much trickier, because the intermediate frame content is designed by the visual artist, with possibly time distortions involved. To illustrate this, in Fig.~\ref{fig:distro-tau}, we depict the distribution of estimated values of $\tau$ for triplets of consecutive images in the case of natural videos (left) and hand-drawn animation videos (right); the value of $\tau$ is estimated by an image-based algorithm that we will describe in~\ref{subsec:tau}. As it can be seen, the distribution of $\tau$ has a much larger variance in the case of hand-drawn animation triplets. This motivates us in re-estimating $\tau$ for all the triplets used for our model training process. 

Our model performs a diffusion-like transition from the target image $\mathbf{I}_{\tau}$ to a noisy version of the conditional images $\mathbf{I}_0$ and $\mathbf{I}_1$. Following the state of the art works on diffusion~\cite{Nonequilibrium_Thermodynamics,DDPM_original,ScoreBased_original}, we model the diffusion process as follows

\begin{equation}
\label{eq:difffusion inference}
p\left(\mathbf{I}_{\tau} \mid \mathbf{I}_0, \mathbf{I}_1, {\tau}\right)=\int p\left(\mathbf{I}_{\tau}^{(T)} \mid \mathbf{I}_0, \mathbf{I}_1, {\tau}\right) \prod_{t=1}^Tp_{\boldsymbol{\theta}}(\mathbf{I}_{\tau}^{(t - 1)} \mid \mathbf{I}_{\tau}^{(t)}, \mathbf{I}_0, \mathbf{I}_1, {\tau}) \mathrm{d} \mathbf{I}_{\tau}^{(1:T)},
\end{equation}
where $t=1,...,T$ are the diffusion steps and $p_{\theta}$ describes the probability of obtaining a partially denoised intermediate frame $\mathbf{I}^{(t-1)}_{\tau}$ at timestep $t-1$, given the boundary frames $\mathbf{I}_0,\mathbf{I}_1$ and $\mathbf{I}^{(t)}_{\tau}$. To do so, we approximate this distribution as $p_{\theta} \sim \mathcal{N}(\mu_{\boldsymbol{\theta}}(\mathbf{I}_{\tau}^{(t)}, \mathbf{I}_0, \mathbf{I}_1, {\tau}), \sigma\boldsymbol{\mathrm{I}})$~\cite{DDPM_original}, with $\mu_{\boldsymbol{\theta}}$ implemented as a neural network. This scheme closely aligns with the vision of the ResShift diffusion model~\cite{Resshift_original}, originally implemented in the context of super-resolution, handling only a single conditional image. In our case, we have to feed two conditional inputs ($\mathbf{I}_0$ and $\mathbf{I}_1$) instead of just one. In this work, we initially experimented with the ResShift diffusion model using a unique conditioning formulation defined as $\alpha \cdot \mathbf{I}_0 + (1 - \alpha) \cdot \mathbf{I}_1$ with $\alpha \in [0, 1]$. However, this approach did not perform as well as we expected. Furthermore, we seek a diffusion model that fully incorporates all conditioning information in both the forward and backward processes. This raises a question: Can we design a conditional diffusion model using $n > 1$ conditions such as the one used in ResShift \cite{Resshift_original}?  We discuss a potential answer to this question in the following section.


\subsection{Multiple-Input Residual Diffusion: Forward process}


To generalize the ResShift diffusion models~\cite{Resshift_original} for $n > 1$ inputs or conditions, we consider $\tau \in [0,1]$ and the \emph{residuals} between a frame $\mathbf{I}_\tau$ and a set of condition images $\mathbf{J}_i \in \mathcal{J} \triangleq \{\mathbf{J}_1, ..., \mathbf{J}_n\}$ as $R_i(\mathbf{I}_\tau) \triangleq \mathbf{J}_i - \mathbf{I}_\tau$. The case $n=1$ is the one of ResShift, with $R_1$ taken between the high resolution $\mathbf{I}_\tau$ and low resolution $\mathbf{J}_1$ images. Keeping the premises of the shifting sequence from the original paper, we define a new one $\{\sum_{i=1}^n\eta_i^{(t)}\}_{t=1}^T$, which increases monotonically with the time steps and satisfies $\sum_{i=1}^n\eta_i^{(1)} \to 0$ and $\sum_{i=1}^n\eta_i^{(T)} \to 1$. For a given $t$, the values $\eta_i^{(t)},\; \forall i \in [1, n]$ can variate as needed (we come back to that point in~\ref{sec:noise schedule}). The Gaussian transition distribution is formulated as

\begin{equation}
\label{eq:forward distribution}
q\left(\mathbf{I}_\tau^{(t)} \mid \mathbf{I}_\tau^{(t-1)}, \mathcal{J}\right)=\mathcal{N}\left(\mathbf{I}_\tau^{(t)} ; \mathbf{I}_\tau^{(t-1)}+\sum_{i=1}^n \alpha_i^{(t)} R_i(\mathbf{I}_\tau), \kappa^2 \sum_{i=1}^n \alpha_i^{(t)} \boldsymbol{\mathrm{I}}\right), t=1,2, \cdots, T
\end{equation}
where $\alpha_i^{(t)}\triangleq\eta_i^{(t)}-\eta_i^{(t-1)}$ for $t>1$ and $\alpha^{(1)}_i\triangleq\eta^{(1)}_i, \kappa$ is a scaling hyper-parameter that controls the noise variance. We can sample data from this distribution through 
\begin{equation}
\label{eq:one-step-marginal}
\mathbf{I}_\tau^{(t)}=\mathbf{I}_\tau^{(t-1)}+\sum_{i=1}^n \alpha_i^{(t)} R_i(\mathbf{I}_\tau)+\kappa \sqrt{\sum_{i=1}^n \alpha_i^{(t)}} \epsilon^{(t)},
\end{equation}
where $\epsilon^{(t)} \sim \mathcal{N}(0, \boldsymbol{\mathrm{I}})$. Given this, we can sample noisy versions of the original image $\mathbf{I}_\tau$, bypassing all future steps until $t \geq 1$, through the following equation (a proof is given in appendix~\ref{apendix:A})  

\begin{equation}
\label{eq:t-step-marginal}
\mathbf{I}_{\tau}^{(t)} = \mathbf{I}_{\tau} + \sum_{i=1}^n \eta^{(t)}_iR_i(\mathbf{I}_\tau )+\kappa \sqrt{\sum_{i=1}^n \eta^{(t)}_i} {\epsilon}.
\end{equation}

Based on this formulation, one sees that when $t\to T$, the Gaussian mean tends to $\sum_{i=1}^n \eta^{(T)}_i \mathbf{J}_i$, i.e. a weighted (with the weights summing to one) combination of the $\mathbf{J}_i$. Similarly, when $t \to 0$ the mean tends to the ground truth $\mathbf{I}_{\tau}$. Hence, the marginal distributions of $\mathbf{I}_{\tau}^{(1)}$ and $\mathbf{I}_{\tau}^{(T)}$ converge to $\delta_{(\mathbf{I}_{\tau}^{(0)})}(.)$ and $\mathcal{N}(\sum_{i=1}^n \eta^{(T)}_i \mathbf J_i, \kappa^2  \mathbf{I})$, respectively, just as happens in the original ResShift model. 

The variance of the diffusion process is given by $\kappa^2 \sum_{i=1}^n \eta^{(t)}_i \mathbf{I}$, indicating that the dispersion of the image increases with the diffusion steps. In the limit $t \to T$, the variance reaches its maximum value $\kappa^2 \mathbf{I}$, meaning that the image is distributed around the mean $\sum_{i=1}^n \eta^{(T)}_i \mathbf{J}_i$ with uncertainty controlled by $\kappa$. On the other hand, when $t \to 0$, the variance tends to zero, ensuring that the initial image $\mathbf{I}_{\tau}$ remains practically unchanged at the beginning of the diffusion process.

\subsection{Multiple-Input Residual Diffusion: Sampling Process}

Now, let us use the diffusion framework to estimate the distribution over $\mathbf{I}_\tau$ conditioned to $\mathcal{J}$,
\begin{equation}
p\left(\mathbf{I}_{\tau} \mid \mathcal{J}\right) = \int p\left(\mathbf{I}_{\tau}^{(T)} \mid \mathcal{J}\right) \prod_{t=1}^T p_{\boldsymbol{\theta}}\left(\mathbf{I}_{\tau}^{(t-1)} \mid \mathbf{I}_{\tau}^{(t)}, \mathcal{J}\right) \mathrm{d} \mathbf{I}_{\tau}^{(1:T)},
\end{equation}
and given $p\left(\mathbf{I}_{\tau}^{(T)} \mid \mathcal{J}\right) \approx \mathcal{N}\left(\sum_{i=1}^n \eta^{(T)}_i \mathbf{J}_i, \kappa^2 \boldsymbol{\mathrm{I}}\right)$, we can suppose that 
\begin{equation}
p_{\boldsymbol{\theta}}\left(\mathbf{I}_{\tau}^{(t-1)} \mid \mathbf{I}_{\tau}^{(t)}, \mathcal{J}\right) = \mathcal{N}\left(\boldsymbol{\mu}_{\boldsymbol{\theta}}\left(\mathbf{I}_{\tau}^{(t)}, \mathcal{J}, t\right), \boldsymbol{\Sigma}_{\boldsymbol{\theta}}\left(\mathbf{I}_{\tau}^{(t)}, \mathcal{J}, t\right)\right). 
\end{equation}
Optimizing for $\boldsymbol{\theta}$ is achieved by minimizing the negative evidence lower bound, that is equivalent to minimizing the Kullback-Leibler (KL) divergence~\cite{DDPM_original}  between $q\left(\mathbf{I}_{\tau}^{(t-1)} \mid \mathbf{I}_{\tau}^{(t)}, \mathbf{I}_{\tau}, \mathcal{J}\right)$ and $ p_{\boldsymbol{\theta}}\left(\mathbf{I}_{\tau}^{(t-1)} \mid \mathbf{I}_{\tau}^{(t)}, \mathcal{J}\right)$. To estimate the explicit form of $q\left(\mathbf{I}_{\tau}^{(t-1)} \mid \mathbf{I}_{\tau}^{(t)}, \mathbf{I}_{\tau}, \mathcal{J}\right)$, we use the Bayes rule to factorize the distribution into known distributions, as in the diffusion literature, with 

\begin{equation}
q\left(\mathbf{I}_{\tau}^{(t-1)} \mid \mathbf{I}_{\tau}^{(t)}, \mathbf{I}_{\tau}, \mathcal{J}\right) \propto q\left(\mathbf{I}_{\tau}^{(t)} \mid \mathbf{I}_{\tau}^{(t-1)}, \mathcal{J}\right) q\left(\mathbf{I}_{\tau}^{(t-1)} \mid \mathbf{I}_{\tau}, \mathcal{J}\right).
\end{equation}

Now, by using equations \ref{eq:one-step-marginal} and \ref{eq:t-step-marginal} in the Gaussian formula and applying the logarithm to simplify the exponents, we obtain the following result (see the full developments in Appendix~\ref{apendix:A})

\begin{equation}
\sigma_t^2 =\kappa^2 \frac{(\sum_{i=1}^n\eta_i^{(t-1)})(\sum_{i=1}^n\alpha_i^{(t)})}{\sum_{i=1}^n\eta_i^{(t)}}.
\end{equation}

\begin{equation}
\mu_t=\frac{\sum_{i=1}^n\eta_i^{(t-1)}}{\sum_{i=1}^n\eta_i^{(t)}} \left(\mathbf{I}_\tau^{(t)} 
 +  \sum_{i=1}^n \eta_i^{(t)} R_i(\mathbf{I}_\tau) \right)+\frac{\sum_{i=1}^n\alpha_i^{(t)}} {\sum_{i=1}^n\eta_i^{(t)}} \mathbf{I}_\tau - \sum_{i=1}^n \eta_i^{(t-1)} R_i(\mathbf{I}_\tau).
\end{equation}

Since $R_i$ depends on $\mathbf{I}_\tau$ and $\mathbf{J}_i$, we can simplify the loss function in the same way as the original ResShift diffusion~\cite{Resshift_original}, trying to estimate $\mathbf{I}_{\tau} \approx f_{\boldsymbol{\theta}}\left(\mathbf{I}_\tau^{(t)}, \mathcal{J}, \tau, t\right)$ at all timesteps $t$ through

\begin{equation}
\min _{\boldsymbol{\theta}} \sum_t\left\|f_{\boldsymbol{\theta}}\left(\mathbf{I}_\tau^{(t)}, \mathcal{J}, \tau, t\right)-\mathbf{I}_\tau\right\|_2^2.
\end{equation}

\subsection{Multiple-Input Residual Diffusion: Noise Schedule}
\label{sec:noise schedule}

For the multiple input noise scheduler, we define the progress of $\sqrt{\sum_{i = 1}^n\eta_i^{(t)}}$ instead of $\sum_{i = 1}^n\eta_i^{(t)}$, and all are defined as the original ResShift Diffusion~\cite{Resshift_original}. Given that $\sum_{i = 1}^n\eta_i^{(1)} \rightarrow 0$, we propose

\begin{equation}\sum_{i = 1}^n\eta_i^{(1)} \triangleq \min\left((0.04 / \kappa)^2,0.001\right).
\end{equation}

For $t \in[2, T-1]$, the noise standard deviation is determined through a non-uniform sequence:

\begin{equation}
\sqrt{\sum_{i = 1}^n\eta_i^{(t)}}=\sqrt{\sum_{i = 1}^n\eta_i^{(1)}} \times b_0^{\beta_t}, t=2, \cdots, T-1,
\end{equation}

where

$$
\beta_t=\left(\frac{t-1}{T-1}\right)^p \times(T-1), b_0=\exp \left[\frac{1}{2(T-1)} \log \frac{\sum_{i = 1}^n\eta_i^{(T)}}{\sum_{i = 1}^n\eta_i^{(1)}}\right] .
$$

Knowing $\sqrt{\sum_{i = 1}^n\eta_i^{(t)}}$ we can obtain $\sum_{i = 1}^n\eta_i^{(t)}$ by elevating it to the square. In order to get the individual $\eta_i^{(t)}$, one can define a weight partition $( a_1, ..., a_n)$ where $\sum_{i = 1}^na_i = 1$ that allows to weight every condition $\mathbf{J}_i$. Given that, one can calculate $\eta_k^{(t)} = a_k \cdot\sum_{i = 1}^n\eta_i^{(t)}$ for any $t$. 

For our problem in VFI we have two inputs ($n = 2$) and, for any independent inputs $\mathbf{I}_0, \mathbf{I}_1$, we can choose a partition based on an estimate of which side the middle frame is closer to, as $(\tau, 1 - \tau)$, $\tau \in [0, 1]$. We will get a deeper insight on how to estimate this partition in the next section.

\subsection{Tau Inter Frame Distance}
\label{subsec:tau}
In our approach, we have $\mathcal{J}=\{\mathbf{I}_0, \mathbf{I}_1\}$ and want to estimate the distribution $p\left(\mathbf{I}_{\tau} \mid \mathbf{I}_0, \mathbf{I}_1, \tau\right)$, where $\tau$ is the temporal position of the target frame between the input frames $\mathbf{I}_0$ and $\mathbf{I}_1$. However, during training, we lack access to the ground-truth values of $\tau$. Ideally, we would estimate this temporal information for all triplets of images in our dataset $\mathcal{D} = \{\mathbf{I}_0^{(s)}, \mathbf{I}_{\tau}^{(s)}, \mathbf{I}_1^{(s)}\}_{s \in S}$ with $S$ the set of image indices within the dataset. This approach would allow the model to learn the temporal relationships between frames and to accurately interpolate frames at \emph{arbitrary} positions during inference.

Hence, in this work, we introduce the metric $\tau_{IFD}$. Its logic is to generate a value that describes the amount of movement from $\mathbf{I}_0 \rightarrow \mathbf{I}_{\tau}$ and from $\mathbf{I}_1 \rightarrow \mathbf{I}_\tau$ in critical areas, i.e., areas with significant movement and change. To achieve this, we first identify these areas with a morphological operation of thresholded opening between the differences of the grayscale images. The threshold is calculated by the Otsu adaptive algorithm~\cite{otsu_threshold_original}. We apply the morphological operation before thresholding to preserve the areas with large structures. Let us define $\Delta \mathbf{I}_{0 \rightarrow {\tau}} = \mathbf{I}^{gray}_{\tau} - \mathbf{I}^{gray}_0$ and $\Delta \mathbf{I}_{1 \rightarrow {\tau}} = \mathbf{I}^{gray}_{\tau} - \mathbf{I}^{gray}_1$.

We define $\mathbf{B}$ as windows filled with ones, of size $k = 5$, and

\begin{equation}
     \mathbf{M}_{0 \rightarrow {\tau}} = \left| \Delta \mathbf{I}_{0 \rightarrow {\tau}} \circ \mathbf{B}\right| > \delta, \quad   \mathbf{M}_{1 \rightarrow {\tau}} = \left| \Delta \mathbf{I}_{1 \rightarrow {\tau}} \circ \mathbf{B} \right| > \delta,
\end{equation}
with $\delta$ a threshold. Then, the optical flow between the two pairs of images is estimated to detect the amount of movement. In this case, the RAFT model~\cite{RAFT_original} is used, which is a recurrent network that processes the feature pyramid correlations between these two images. Finally, since the information needed is only the amount of movement, the magnitude of the optical flow is calculated as follows

\begin{equation}
     \mathbf{F}_{0 \rightarrow {\tau}} = \text{RAFT}(\mathbf{I}_0, \mathbf{I}_{\tau}), \quad ||\mathbf{F}_{0 \rightarrow {\tau}}|| = \sqrt{\mathbf{F}_{0 \rightarrow {\tau}, x}^2 + \mathbf{F}_{0 \rightarrow {\tau}, y}^2},
     \label{eq:f0n}
\end{equation}

\begin{equation}
    \mathbf{F}_{1 \rightarrow {\tau}} = \text{RAFT}(\mathbf{I}_1, \mathbf{I}_{\tau}), \quad ||\mathbf{F}_{1 \rightarrow {\tau}}|| = \sqrt{\mathbf{F}_{1 \rightarrow {\tau}, x}^2 + \mathbf{F}_{1 \rightarrow {\tau}, y}^2}.
    \label{eq:f1n}
\end{equation}

Finally, the sum of all the magnitudes of movement considered within a critical zone is calculated, and then they are normalized to fall within the range of $(0, 1)$, leading to


\begin{equation}
    \tau_{IFD} = \frac{\sum ||\mathbf{F}_{0 \rightarrow {\tau}}|| \cdot \mathbf{M}_{0 \rightarrow {\tau}}}{\sum ||\mathbf{F}_{1 \rightarrow {\tau}}|| \cdot \mathbf{M}_{1 \rightarrow {\tau}} + \sum ||\mathbf{F}_{0 \rightarrow {\tau}}|| \cdot \mathbf{M}_{0 \rightarrow {\tau}} }.
    \label{eq:alpha_ifd}
\end{equation}

This estimate is used \emph{at training times} to feed the interpolation module.

\section{Description of the deep Learning Framework}

\begin{figure}[t]		
    \centering
    \includegraphics[scale = 0.30]{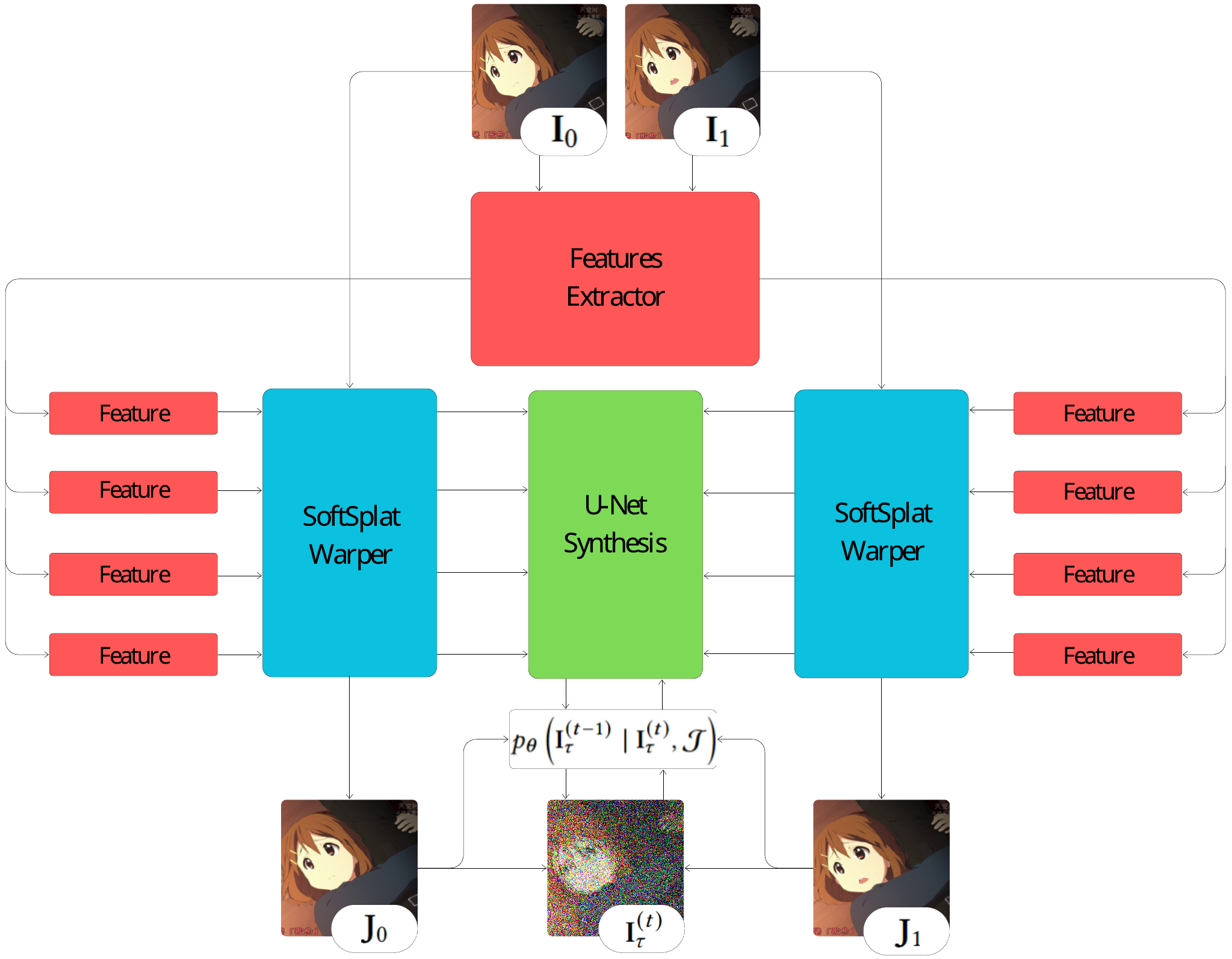}	
    \caption{General overview of the proposed model.}
    \label{fig:overview}
\end{figure}

Given the formulation of our Multiple-Input Residual Diffusion described above, we have to deal with two inputs: The initial image $\mathbf I_0$ and the final image $\mathbf I_1$. Hence, our denoiser module for the residual diffusion process has to be aware of these inputs. At each stage of the Markov chain induced by the Multiple-Input Residual Diffusion model, an estimate of the target image $\mathbf I_{\tau}$ is generated. Our idea is to combine deep learning methods with a forward warping technique such as Softmax splatting, following a strategy similar to that proposed in~\cite{SoftmaxSplating_original}, and successfully applied in other VFI studies related to animation~\cite{AnimeInterp_original,EiseiAnimeInterp_original}. In \ref{fig:overview}, the overall model structure is depicted, with its three main stages: Feature Extraction, Feature Warping, and Image Synthesis. 

Our model first takes the input images—the initial and final frames—and extracts multi-scale features from each, including edges, that we pass explicitly to the model. Simultaneously, it estimates the optical flow in both directions, $\mathbf F_{0 \rightarrow 1}$, $\mathbf F_{1 \rightarrow 0}$. Using these inputs and optical flows, a warping process is performed with Softmax Splatting, applying it to both the features and the images in both forward and backward directions. Finally, the warped representations are processed by a UNet, acting as a synthesizer, refining the results by correcting artifacts introduced during warping, such as blurring and gaps. These components are explained in greater detail in the following sections.

\subsection{Feature Extraction}
\label{sec:feature_extraction}
The feature extraction module follows a hierarchical architecture with four progressive reduction levels, where the number of channels increases from 128 to 256, then to 512, and remains at 512 in the final stage. Each level consists of two consecutive 2D convolutions with a $3 \times 3$ kernel. The first convolution applies downsampling with a stride of 2 to capture multi-scale features efficiently, while the second refines the extracted representations. SiLU activation is used to introduce non-linearity, and Group Normalization is incorporated to enhance generalization and mitigate overfitting.


In the realm of traditional animation, edge preservation is a fundamental aspect, as even slight blurring can significantly degrade the visual quality of an image. To mitigate this effect, and based on the positive results obtained from incorporating an explicit edge detection approach in~\cite{EiseiAnimeInterp_original}, we propose to use of Difference of Gaussians (DoG) as an explicit edge detection method,

\begin{equation}
    DoG(\mathbf I)=\frac{1}{2}+k_t\cdot\left(G_{k_\sigma \sigma}(\mathbf I)-G_\sigma(\mathbf I)\right)
\end{equation}
where $G_{\sigma}$ are Gaussian blurs, $k_\sigma = 1.6$ and $k_t = 2$. Next, an Euclidean Distance Transform (EDT) is applied to a thresholded DoG at 0.5, which represents the proximity of each pixel to the detected edges. To ensure that the resulting values remain within a bounded range and facilitate their integration into the model architecture, the EDT values are normalized to a unit range,

\begin{equation}
    NEDT(\mathbf I)=1-\exp \left[\frac{-EDT(DoG(\mathbf I)>0.5)}{d}\right]
\end{equation}
where $d = 15$ is a steepness hyperparameter.

\subsection{Softmax Splatting Warping}
\label{sec:warping}

\begin{figure}[t]		
    \centering
    \includegraphics[scale = 0.35]{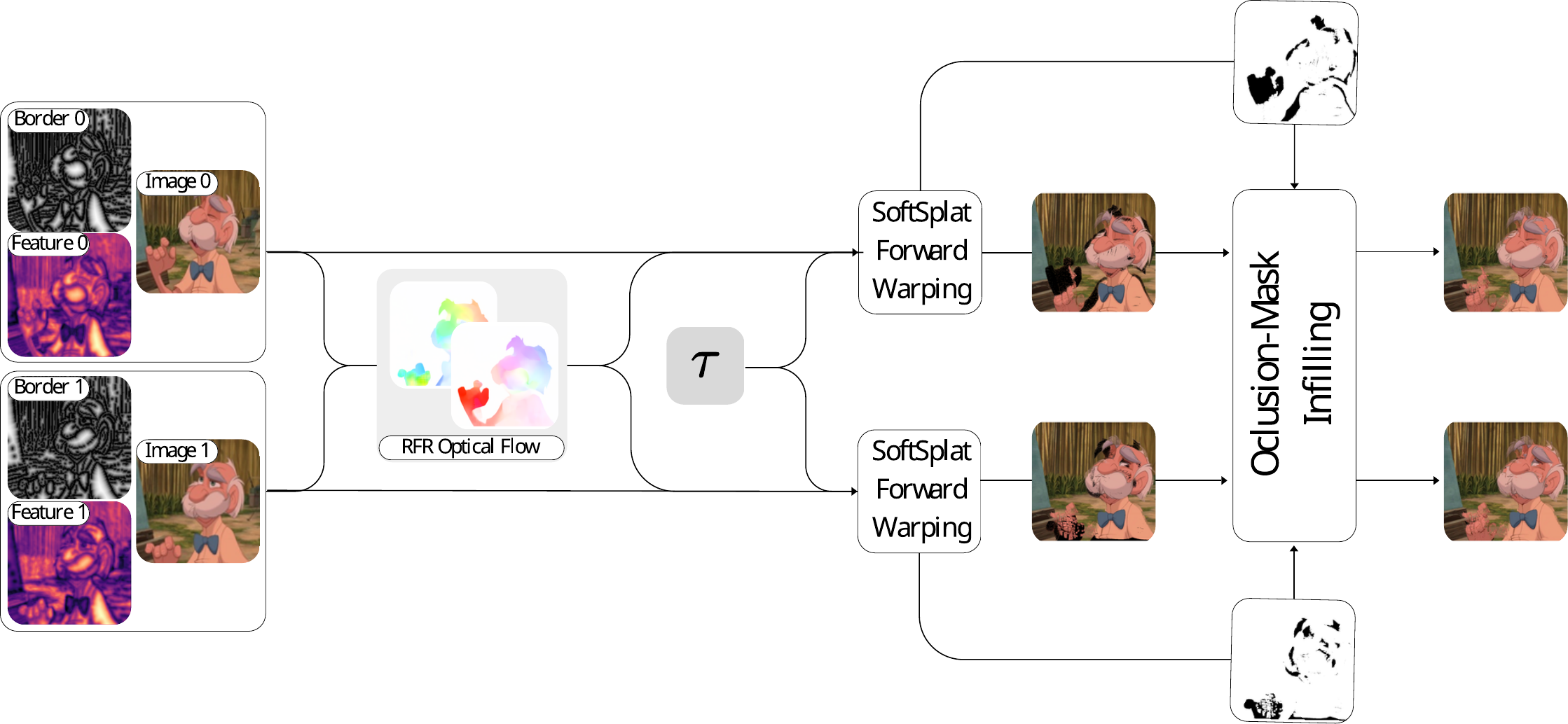}	
    \caption{Initial warping: Based on Softmax Splatting~\cite{SoftmaxSplating_original}, we produce two initial versions of the intermediate images $\mathbf{\hat I}_{0 \rightarrow \tau}$ and $\mathbf{\hat I}_{\tau \rightarrow 1}$.}
    \label{fig:feature_warping}
\end{figure}

In this section, all the used models have their weights frozen. Building upon previous works~\cite{EiseiAnimeInterp_original}, we obtain an initial interpolation by processing the frames in both directions ($\mathbf I_0 \rightarrow \mathbf I_1$ and $\mathbf I_1 \rightarrow \mathbf I_0$) as we can see in figure~\ref{fig:feature_warping}. We perform this operation for RGB images, feature maps and edge maps.

First, optical flows are estimated in these two directions and serve as the basis for forward warping using Softmax Splatting~\cite{SoftmaxSplating_original}. We define the importance metric $Z$ as in~\cite{EiseiAnimeInterp_original}, using the brightness constancy as an occlusion indicator. It is derived through backward warping $\overleftarrow{\omega}$ as 

$$Z=-0.1 \cdot\left\|\mathbf I_0-\overleftarrow{\omega}\left(\mathbf I_1,\mathbf F_{0 \rightarrow 1}\right)\right\|_1.$$

The model used for estimating optical flow is RFR, a modified RAFT model~\cite{RAFT_original} tailored for animated images~\cite{AnimeInterp_original}. 
When performing forward warping, it is common for some images to have empty spaces due to object displacement and the method’s limited ability to infer the missing content in these areas. Hence, we fill these sections using the following formula proposed in~\cite{EiseiAnimeInterp_original} 

\begin{equation}
\label{eq:relleno}
    \begin{aligned}
        \mathbf{\tilde I}_{0 \rightarrow \tau}&=\frac{1}{2}\left(\mathbf M_{0 \rightarrow \tau} 
        \cdot \mathbf{\hat I}_{0 \rightarrow \tau}\cdot \mathbf I_0\right.  \left.+\left(1-\mathbf M_{0 \rightarrow \tau}\right)\cdot \mathbf{\hat I}_{1 \rightarrow \tau}\cdot \mathbf I_1\right)\\
        &+\frac{1}{2}\left(\mathbf M_{1 \rightarrow \tau}\cdot \mathbf{\hat I}_{1 \rightarrow \tau} \cdot \mathbf I_1+\left(1-\mathbf M_{1 \rightarrow \tau}\right)\cdot \mathbf{\hat I}_{0 \rightarrow \tau} \cdot \mathbf I_0\right),
    \end{aligned}    
\end{equation}
where $\mathbf{\hat I}_{a \rightarrow b}$ denotes the forward warping from timestep $a$ to timestep $b$, $\mathbf M$ is the occlusion mask after applying morphological opening, and $\mathbf I$ refers to one of the input images. Essentially, occluded regions are filled using the warped features from the other source image. The mask $\mathbf M$ is computed by warping an image of ones and applying a morphological opening with a kernel size of $k=5$ to remove small dotted artifacts. Note that, while the opening operation is non-differentiable, computing the gradient with respect to the flow field is unnecessary, as the flow estimator is fixed.


\subsection{U-Net Synthesizer}

\begin{figure}[t]		
    \centering
    \includegraphics[scale = 0.35]{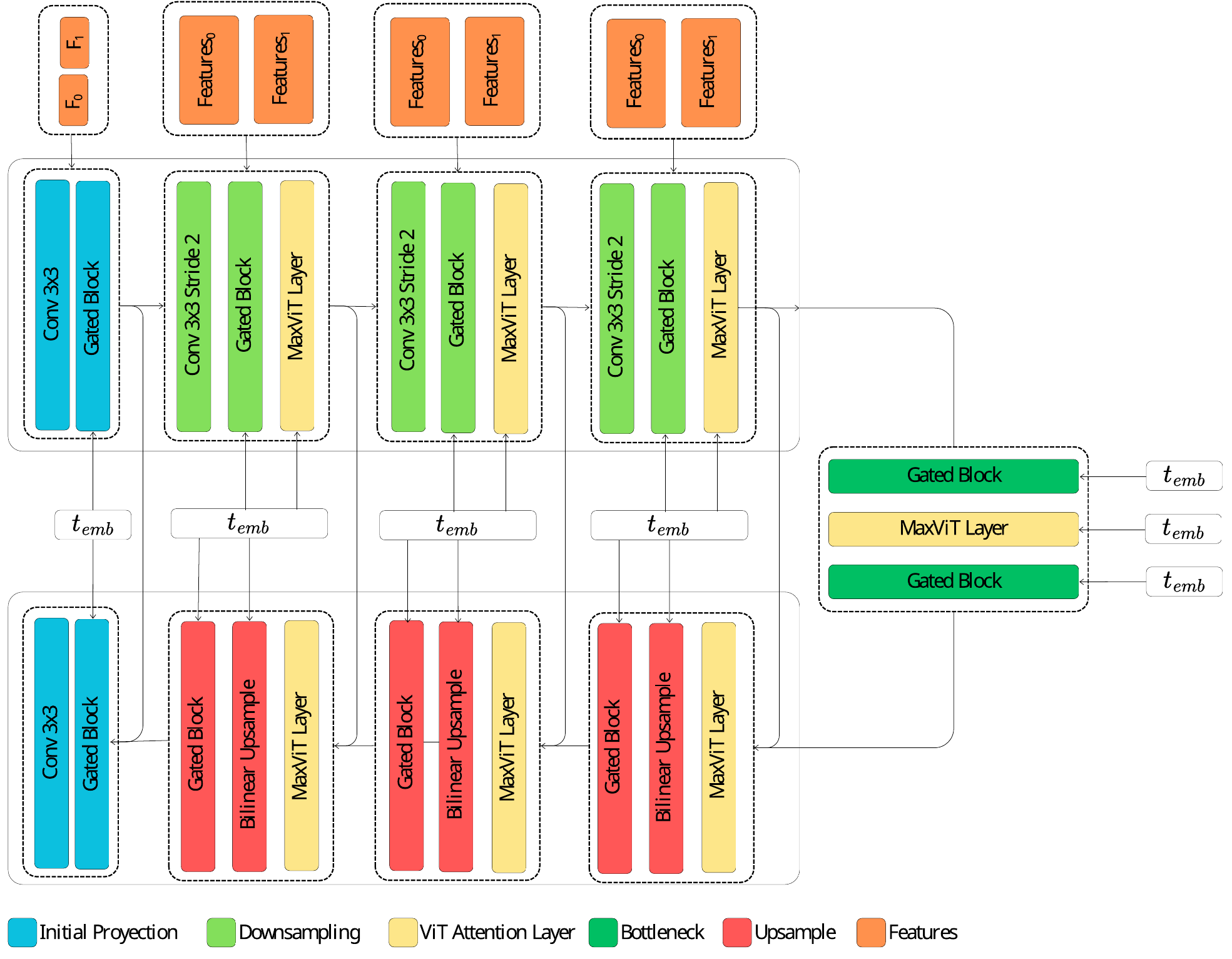}	
    \caption{Architecture of the proposed U-Net based synthesizer. Note that the diffusion timestep $t$ is passed to all the intermediate levels.}
    \label{fig:synthesizer}
\end{figure}

The architecture of the synthesizer is a U-Net with four residual connections. Based on the experiments conducted in~\cite{Resshift_original}, we know that denoisers for this type of model do not need to be massively large to achieve realistic results, unlike other types of diffusion models~\cite{DDIM_original,DDPM_original}. This allows us to create a lightweight and efficient model, as this type of diffusion also does not require many steps in the reverse Markov chain to achieve the desired results. As seen in Figure~\ref{fig:synthesizer}, we pass the corresponding warped features $\mathbf{\tilde I}_{0 \rightarrow \tau},\mathbf{\tilde I}_{\tau \rightarrow 1}$ (orange in Fig.~\ref{fig:synthesizer}) at each encoder stage by concatenating them with the current feature map before feeding it into the bottleneck.



As commented above, the warping process often results in holes or missing regions within the image. These gaps are initially filled using the equation~\ref{eq:relleno}. The binary masks used in that formula indicate the locations of the missing areas, and are then concatenated with their corresponding feature or image tensors, along with the warping borders that provide structural guidance.

To leverage this information, we propose the use of Gated Convolutions ~\cite{Gated_conv-original}, capable of dynamically modulating the importance of the initially inpainted regions. By incorporating the warping borders as a source of guidance, the model can more effectively learn to preserve coherent structures and boundaries during the interpolation process. Gated Convolutions can be described as follows:

\begin{equation}
    \mathbf Y = \sigma(\mathbf W_g * \mathbf X) \odot \phi (\mathbf W_f * \mathbf X),
\end{equation}
where $\sigma$ denotes sigmoid activation, $*$ is the convolution operator, $\odot$ is the element-wise multiplication and $\phi$ an activation function. All the Double Convolutional blocks shown in the figure are composed of two gated convolution layers, each followed by a SiLU activation function, also processing the temporal sinusoidal embedding $t_{emb}$ between convolutional layers.

\begin{figure}[t]		
    \centering
    \includegraphics[scale = 0.30]{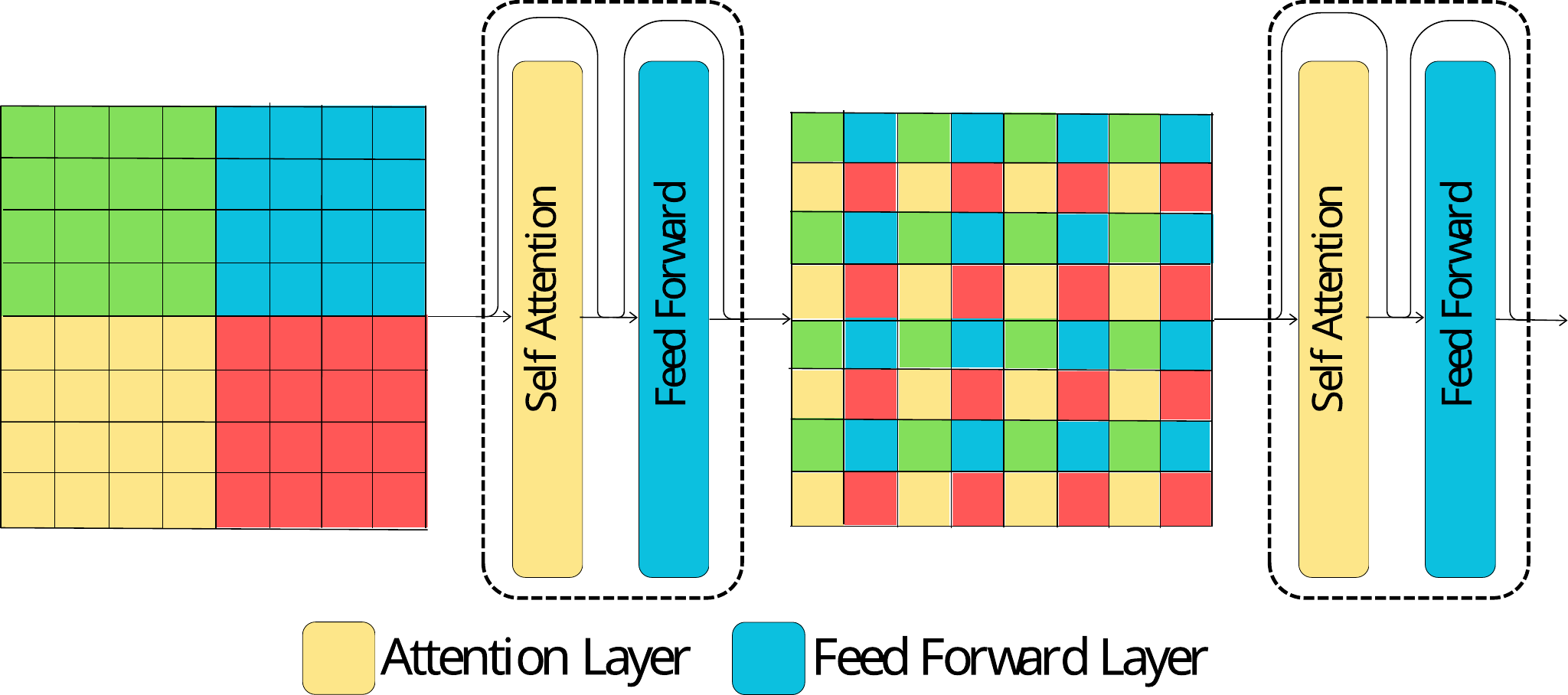}	
    \caption{MultiAxis Transformer Block \cite{MaxViT_original}, Window (local) and grid (global) attention layers. Normalization and activation layers are omitted for simplicity.\label{fig:MAViT}}
\end{figure}

The attention mechanism used in this work (pale yellow in Fig.~\ref{fig:synthesizer}) is the Multi-Axis Self-Attention introduced in~\cite{MaxViT_original}. This block, described in Fig.~\ref{fig:MAViT}, completely decomposes the dense attention mechanisms into two forms: window and grid attention, reducing the quadratic complexity of vanilla attention to linear, without loss of non-locality. This simple and flexible design often performs even better than full attention schemes. In VFI, it is efficiently used in~\cite{LDMVFI_original} to lighten the model and at the same time capture information at two levels of movement. Given $\mathbf X \in \mathbb{R}^{H \times W \times C}$ an input feature map, instead of applying attention on the flattened spatial dimension $H \times W$, we partition the tensor as $(H / P \times W / P, P \times P, C)$, a division into non-overlapping windows, each of size $P \times P$ (left grid of Fig.~\ref{fig:MAViT}). Applying self-attention within the local spatial dimension, i.e., $P \times P$, allows to conduct \emph{local interactions}. Similarly, rather than partitioning feature maps using a fixed window size, we grid the tensor into the shape $(G \times G, H / G \times W / G, C)$ using a fixed $G \times G$ uniform grid, resulting in windows of adaptive size $H / G \times W / G$. Using self-attention on the decomposed grid axis, i.e., $G \times G$, corresponds to dilated, global spatial mixing of tokens, capturing global interactions (Fig.~\ref{fig:MAViT}, right). To incorporate the information from the temporal embedding, this layer employs an Adaptive Layer Normalization, following~\cite{DiT-original}. Moreover, typical design choices from Transformers are adopted, including Feed Forward Networks (FFN)~\cite{ViT_original, Swin_Transformer_original} and residual connections.


\subsection{Loss function}

The composition of the three modules described above (features extraction, warping and synthesizer) is denoted as 
$D_\theta(\mathbf{I_{\tau}^{(t)}}, \mathbf{I}_0, \mathbf{I}_1, \hat{\tau}, t)$: with parameters $\theta$, it processes a noisy image $\mathbf{I_{\tau}^{(t)}}$ corresponding to a diffusion step $t$, conditions $\mathbf{I}_0, \mathbf{I}_1$ and an estimate of the interpolation time $\hat{\tau}$.
As in~\cite{Resshift_original}, the loss function $\mathcal L(\theta)$ used to optimize the parameters $\theta$ of the denoiser measures the difference between the estimate of $\mathbf I_\tau$ from the noisy version of this image, at time-step $t$. Here, we slightly modify this loss function in that we use the sum of a classical MSE term with a LPIPS \cite{LPIPS_original} term that use a vgg16 network, weighted by a factor of $0.2$, as described in the following equation

\begin{equation}
\mathcal L(\theta) \triangleq E_{t,(\mathbf I_0,\mathbf I_\tau,\mathbf I_1)}\left[ \| \mathbf D_\theta(\mathbf{I_{\tau}^{(t)}}, \mathbf{I}_0, \mathbf{I}_1, \hat{\tau}, t) - \mathbf I_\tau\|^2+0.2 \cdot \lpips(\mathbf D_\theta(\mathbf{I_{\tau}^{(t)}}, \mathbf{I}_0, \mathbf{I}_1, \hat{\tau}, t),\mathbf I_\tau)\right].
\end{equation}

\section{Experimental results}
\subsection{Experimental setup}

Within the denoising module presented in Fig.~\ref{fig:synthesizer}, an overall reduction factor of $\times 8$ is taken, where each Downsampling block reduces the size of the image by half. We choose a base channel size as $128$, and multiply this value along the following blocks until we reach $512$ channels. The output dimensions of the sinusoidal positional encoders, both for $\hat{\tau}$ and for the time step $t$, are taken as $512$, as it is the maximum number of channels that the processing input image can reach. 

For the configuration of the sampler or ResShift Diffusion, the noise control factor is taken as $\kappa = 2.0$. Given this aforementioned noise addition factor, $T = 20$ timesteps are taken for the forward/reverse processes. We take the maximum value of the shift sequence as $\eta_K = 0.99$ as recommended in the original paper~\cite{Resshift_original}. The growth factor characterizing the variance schedule $\eta^{(k)}$ is taken as $p = 0.3$. Finally, the minimum noise level $\eta^{(1)}$ is taken as $0.04$.

We implement our model in PyTorch, using Lightning as a complement package only for the training pipline implementation. As commented in~\ref{sec:warping}, we use RFR/RAFT~\cite{RAFT_original, AnimeInterp_original} for the optical flow estimation. We train with the AdamW optimizer \cite{AdamW-original} with a learning rate of $10^{-4}$ using a scheduler strategy that monitors the validation loss after each epoch and reduces the learning rate by a factor of 0.5 if no improvement is observed over a patience window of 3 consecutive epochs. We train our model for 50 epochs with a batch size of 6, and accumulate gradients of 5 for an effective batch size of 30. Also, we clip the gradients in the range $(0, 1)$. All model are trained and tested at a $256 \times 448$ of image resolution. Two NVIDIA Titan GPUs were used for training and evaluation. Upon 

\subsection{Datasets and metrics}

As a training dataset, we mainly employ ATD-12K introduced in~\cite{AnimeInterp_original} and used in other animation VFI works like \cite{EiseiAnimeInterp_original}. Thus, our final training set consists of random 9000 frame triplets $(\mathbf I_0,\mathbf I_\tau,\mathbf I_1)$ from ATD-12K and separate 1000 for validation and 2000 for testing steps. To increase the diversity of the data, we perform random reversals in the temporal order and apply random spatial flips to the triplets. We recall that, as described in Section~\ref{subsec:tau}, an important pre-processing step during training is to estimate $\hat{\tau}$ which is then used to fit the estimated optical flows.

The metrics we use to compare methods include the classical PSNR and SSIM metrics, but our mains metrics are the perceptuals ones, LPIPS~\cite{LPIPS_original} and FloLPIPS. FloLPIPS~\cite{FloLPIPS_original} is an extension of LPIPS designed to measure both perceptual distance and differences in motion using optical flows, and it is particularly useful in video frame interpolation problems. Roughly speaking, FloLPIPS calculates the optical flow before passing the images through the LPIPS calculation and it weights the LPIPS metric with the amount of optical flow.

\subsection{Quantitative results}

We compare our method against various representative state-of-the-art models, namely cartoon video interpolation methods. The baselines we compare our method with are ABME~\cite{ABME}, Anime Interp~\cite{AnimeInterp_original}, Eisei~\cite{EiseiAnimeInterp_original} and Softsplat \cite{SoftmaxSplating_original}. In Table~\ref{tab:sota-compare}, we can see that our method obtains systematically the best results on the aforementioned metrics. In relative numbers, the improvement is even greater for the perceptual metrics (LPIPS and FloLPIPS), which may be partially explained by the fact that our objective function includes an LPIPS term.

\begin{table}[t]
    \centering
    \resizebox{8.5cm}{1.7cm}{%
    \begin{tabular}{lccccr}
         & \multicolumn{4}{c}{ATD-12k-test} \\
        \cline{2-5} 
         & $\downarrow$ LPIPS & $\downarrow$ FloLPIPS & $\uparrow$ SSIM & $\uparrow$ PSNR \\ 
        \hline
        ABME & 0.0424 & 0.112 & 95.19 & 29.07 \\ 
        SoftSplat & 0.0419 & 0.106 & 95.05 & 28.91 \\  
        \hline
        AnimeInterp & 0.0375 & 0.102 & 95.74 & 29.66\\
        Eisei-SSL-DTM & 0.0349 & 0.097 & 95.15 & 29.29\\
        \hline
        Ours & \textbf{0.0322} & \textbf{0.088} & \textbf{95.81} & \textbf{29.82} \\ 
        \hline
    \end{tabular}
    }
    \caption{Quantitative comparison of our model and various methods tested at ATD-12k-test. For each column, the best result is in bold.}
    \label{tab:sota-compare}
\end{table}

In terms of complexity, our model loses some advantage due to its diffusion-based nature, as the sampling implies multiple forward passes to produce the final result. Although our diffusion process requires fewer steps compared to other state-of-the-art methods~\cite{DDIM_original, DDPM_about_1}, as previously stated, it still needs to be sufficiently large to adequately capture the data distribution. Our model consists of 140 million parameters, of which 135M are trainable. The remaining parameters are frozen and correspond to the optical flow RFR model. When compared to other state-of-the-art models, our architecture is significantly larger: Eisei-SSL-DTM (1.28M), AnimeInterp (2.01M), SoftSplat (7.6M), and ABME (17.5M).

\subsection{Qualitative results} 
\label{sec:experimet-qualitative}

\begin{figure}[t]
    \centering
    \begin{subfigure}[b]{\textwidth}
        \centering
        \includegraphics[width=0.99\textwidth]{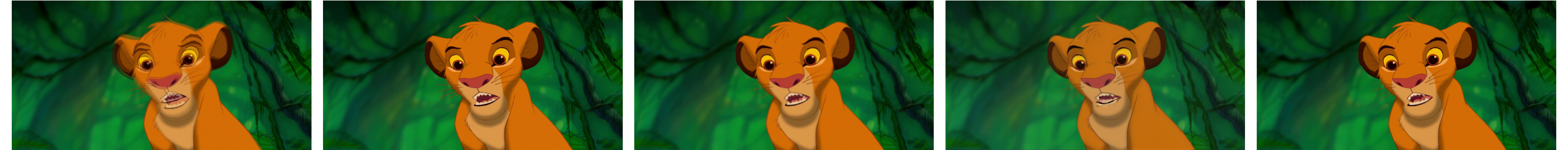}
    \end{subfigure}
    
    \vspace{0.3cm} 

    \begin{subfigure}[b]{\textwidth}
        \centering
        \includegraphics[width=0.99\textwidth]{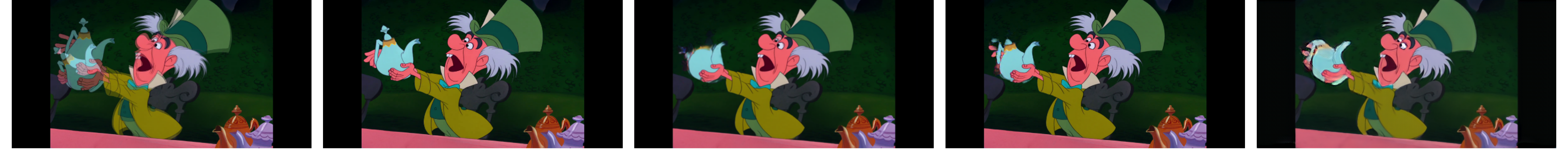}
    \end{subfigure}
    
    \vspace{0.3cm}

    \begin{subfigure}[b]{\textwidth}
        \centering
        \includegraphics[width=0.99\textwidth]{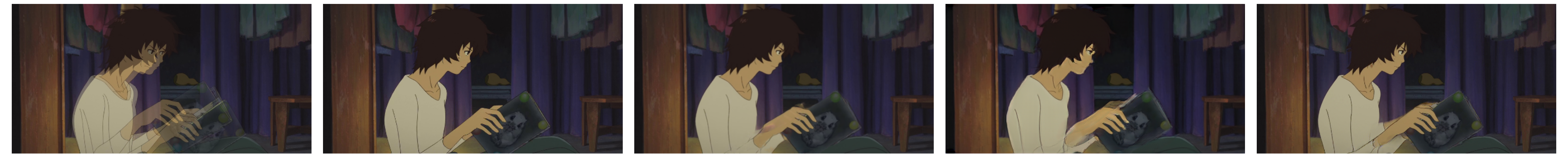}
    \end{subfigure}
    
    \vspace{0.3cm}

    \begin{subfigure}[b]{\textwidth}
        \centering
    \includegraphics[width=0.99\textwidth]{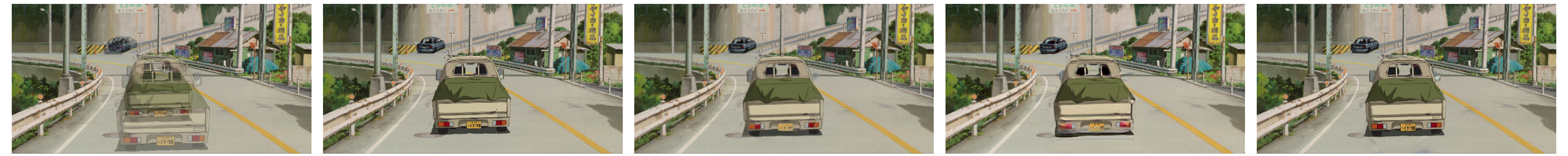}
    \end{subfigure}
    \begin{tabularx}{\textwidth}{YYYYY}
      $\mathbf I_0 + \mathbf I_1$ & GT & Eisei & SoftSplat & Ours 
    \end{tabularx}

    \caption{Qualitative comparison between our method and state-of-the-art (SOTA) video frame interpolation models. Each row corresponds to a different test sequence, and the columns represent (from left to right): frame overlap ($\mathbf I_0 + \mathbf I_1$), ground truth intermediate frame, Eisei, SoftSplat, and our proposed method.}
    \label{fig:sota-qualitative-measurement}
\end{figure}

In figure~\ref{fig:sota-qualitative-measurement}, each row shows a visual comparison for a specific sequence, ordered from left to right as: the simple frame overlay ($\mathbf I_0 + \mathbf I_1$), the ground truth intermediate frame, the results from Eisei and SoftSplat, and our proposed diffusion-based method. Across the examples, Eisei often suffers from blurry or oversmoothed results, particularly noticeable along object edges (e.g., the character's face in row 1 and the tea cup in row 2). SoftSplat provides sharper predictions, but frequently exhibits warping artifacts and visible distortions under motion (row 4). In contrast, our method consistently produces temporally and spatially coherent frames, with well-preserved structure and minimal ghosting. Notably, in challenging regions with occlusion or fast motion (e.g., hand and book interaction in row 3, or car motion in row 4), our approach better approximates the ground truth while maintaining global consistency.

\subsection{Uncertainty analysis}
Given the stochastic nature of diffusion models and their capacity in exploring a full predictive distribution, our case offers an opportunity to study the uncertainty estimated by the model, by generating $N_S$ samples of the interpolated image instead of just one. In the following, we take $N_S=10$. To analyze the corresponding results, we divide the dataset into two distinct domains: Japanese anime and Disney animation. As discussed in Section~\ref{sec:experimet-qualitative}, the model exhibits slightly different behaviors across these domains—an expected outcome given the visual and stylistic differences between them. We then evaluate uncertainty using four metrics: LPIPS, Correlation (CORR), Dynamic Range (MIN-MAX), and the standard deviation (SD) among the $N_S$ samples. 

\begin{table}[t]
    \centering
    \resizebox{8.5cm}{1.0cm}{%
    \begin{tabular}{lccccr}
         & \multicolumn{4}{c}{Variation Metrics} \\
        \cline{2-5} 
         & $\downarrow$ LPIPS & $\uparrow$ CORR & $\downarrow$ MIN-MAX & $\downarrow$ SD \\ 
        \hline
        Disney & 0.0140 & 0.9919 & 0.0299 & 0.0126 \\ 
        Anime & 0.0034 & 0.9979 & 0.0087 & 0.0033 \\  
        \hline
    \end{tabular}
    }
    \caption{Quantitative measurement of uncertainty across two animation domains using the proposed Multi-Input-Resshift-Diffusion model. The Disney subset exhibits higher uncertainty, as indicated by increased LPIPS, standard deviation (SD), and pixel-wise dynamic range (MIN-MAX), along with lower inter-sample correlation. In contrast, the Anime subset shows significantly lower variability, suggesting more confident and consistent predictions.}
    \label{tab:uncertainty-quantitative-measurement}
\end{table}

\begin{figure}[t]
    \centering
    \begin{subfigure}[b]{\textwidth}
        \centering
        \includegraphics[width=0.99\textwidth]{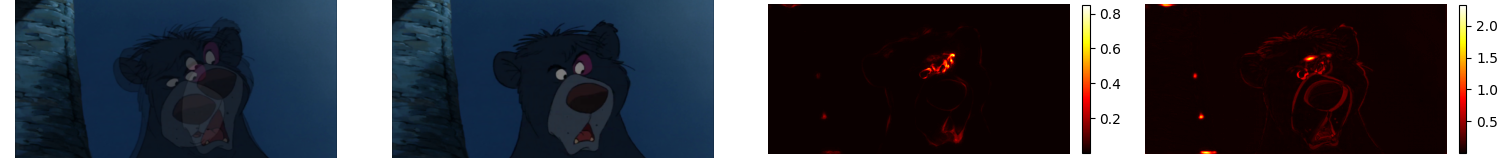}
    \end{subfigure}
    
    \vspace{0.3cm} 

    \begin{subfigure}[b]{\textwidth}
        \centering
        \includegraphics[width=0.99\textwidth]{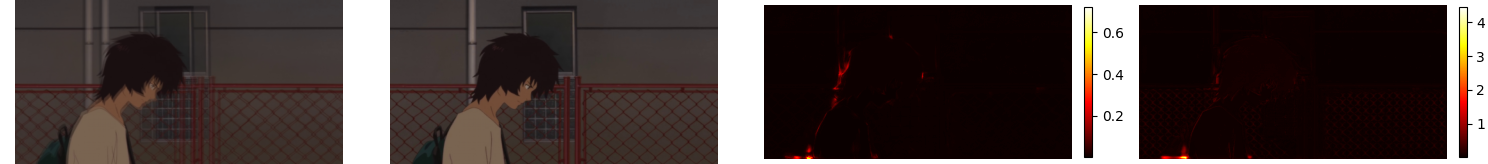}
    \end{subfigure}
    
    \vspace{0.3cm}

    \begin{subfigure}[b]{\textwidth}
        \centering
        \includegraphics[width=0.99\textwidth]{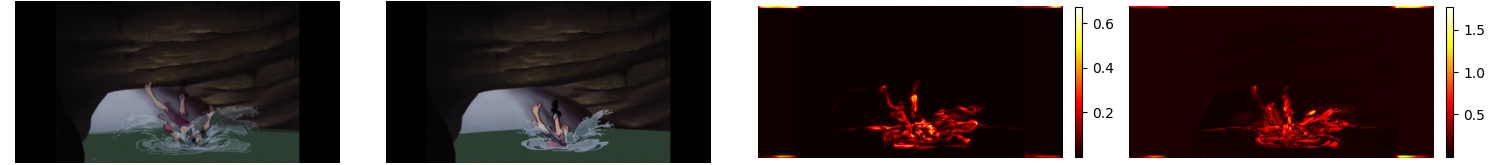}
    \end{subfigure}
    
    \vspace{0.3cm}

    \begin{subfigure}[b]{\textwidth}
        \centering
        \includegraphics[width=0.99\textwidth]{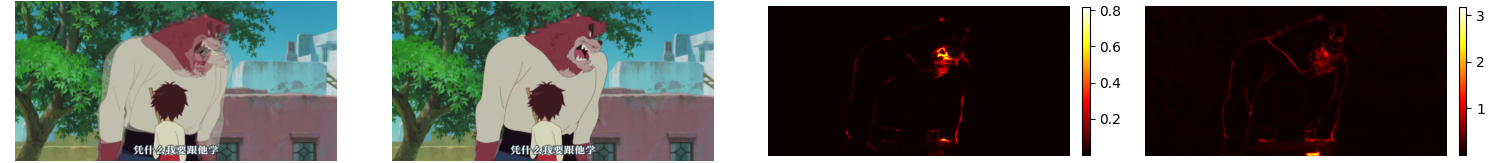}
    \end{subfigure}
    \begin{tabularx}{\textwidth}{YYYY}
      $\mathbf I_0 + \mathbf I_1$ & GT & Deviation Map & RMSE 
    \end{tabularx}
    \caption{Qualitative examples of uncertainty estimated by the proposed VFI diffusion model. Each row shows, from left to right: the mean of the initial and final frames ($\mathbf I_0$ and $\mathbf I_1$), the ground truth central frame $I_\tau$, the pixel-wise standard deviation map computed from $N_S=10$ samples generated by the model conditioned on $\mathbf I_0$ and $\mathbf I_1$, and the RMSE error between the interpolated image and the GT. Regions with higher variability indicate areas where the model exhibits greater uncertainty, usually associated with motion, occlusion, or temporal ambiguity. The uncertainty maps are highly correlated with the error maps.}
    \label{fig:uncertainty-qualitative-measurement}
\end{figure}

As shown in Table~\ref{tab:uncertainty-quantitative-measurement}, the model exhibits higher uncertainty when interpolating frames in Disney animations compared to Japanese anime. This is evidenced by higher LPIPS, standard deviation, and pixel-wise dynamic range, along with slightly lower inter-sample correlation. These results suggest that the model finds Disney sequences more ambiguous or complex, possibly due to their more detailed textures or smoother gradients compared to the typically flatter and stylized anime frames. One possible explanation is the greater amount of global motion typically present in the Disney subset of our dataset, in contrast to the more static compositions often found in traditional Japanese anime, where camera movement is minimal and character motion tends to be discrete, stylized, and spatially concentrated—for example, during dialogue scenes, often only the mouth of a character moves while the rest of the frame remains nearly static.

As shown in figure~\ref{fig:uncertainty-qualitative-measurement}, the pixel-wise uncertainty is quantified using the standard deviation computed across $N_S=10$ interpolated samples generated by the model, conditioned on the same input frames ($\mathbf I_0$, $\mathbf I_1$). The resulting standard deviation maps highlight regions where the model shows less confidence in its predictions. For instance, higher variability is observed in areas involving motion (e.g., the bear's face in the Disney example, or the character's arm in the final row), occlusions (as in the water splash), or ambiguous temporal transitions (e.g., hair movement in the anime sequences). In contrast, background regions or static areas consistently show low standard deviation, indicating high model certainty. Finally, on the rightmost column of figure~\ref{fig:uncertainty-qualitative-measurement}, we also depict the corresponding RMSE errors with respect to the ground truth intermediate images, and one can see that they are rather well correlated with the uncertainty maps.

\begin{table}[t]
    \centering
    \resizebox{8.5cm}{1.2cm}{%
    \begin{tabular}{lccccr}
         & \multicolumn{4}{c}{Variation Metrics} \\
        \cline{2-5} 
         & $\downarrow$ LPIPS & $\uparrow$ CORR & $\downarrow$ MIN-MAX & $\downarrow$ SD \\ 
        \hline
        $\tau = 0.1$ & 0.1285 & 0.9072 & 0.0868 & 0.0372 \\ 
        $\tau = 0.5$ & 0.0153 & 0.9881 & 0.0198 & 0.0081 \\ 
        $\tau = 0.75$ & 0.0183 & 0.9715 & 0.0248 & 0.0103 \\ 
        \hline
    \end{tabular}
    }
    \caption{Quantitative uncertainty analysis  at different interpolation positions $\tau$ using the proposed diffusion model. As $\tau$ moves away from the midpoint (0.5), the model exhibits higher uncertainty, reflected by increased LPIPS, pixel-wise standard deviation (SD), and dynamic range (MIN-MAX), along with decreased correlation between samples. These results indicate that interpolating frames near the temporal boundaries ($\tau = 0.1$ or $0.75$) is inherently more uncertain than interpolating at the center.}
    \label{tab:uncertainty-taus}
\end{table}

To assess how the model's uncertainty varies depending on the temporal position of the interpolated frame, we evaluate the model at different $\tau$ values: 0.1 (closer to $\mathbf I_0$), 0.5 (center), and 0.75 (closer to $\mathbf I_1$). The results, shown in Table~\ref{tab:uncertainty-taus}, reveal that uncertainty is higher when $\tau$ is near the boundaries (e.g., $\tau = 0.1$), and lower at the midpoint ($\tau = 0.5$). Specifically, LPIPS and SD are significantly higher at $\tau = 0.1$, indicating greater variability in generated samples. The correlation between samples also drops from 0.9881 at $\tau = 0.5$ to 0.9072 at $\tau = 0.1$, reinforcing this observation.

These findings align with the training data distributions (Figure~\ref{fig:distro-tau}), where the learned $\tau$ values for photorealistic images are sharply centered around $0.5$, while animated images exhibit a broader distribution. The higher concentration of training samples near $\tau = 0.5$ likely helps the model generalize more confidently to such positions, while extrapolation toward earlier or later frames introduces greater ambiguity and stochasticity.

\subsection{Ablation experiments}

To better understand the contribution of each component in our proposed architecture, we conducted an ablation study by incrementally enabling or disabling key modules. We evaluated six model variants on 300 randomly selected samples from the ATD-12k-test dataset, using LPIPS and PSNR as metrics to quantify perceptual quality and pixel-wise accuracy, respectively.

In table~\ref{tab:ablation}, the V1 variant disables the synthesizer module while preserving the rest of the pipeline, leading to the worst LPIPS score, indicating its critical role in preserving perceptual consistency. V2 re-enables the synthesizer but removes the initial infill, showing moderate improvements in LPIPS and PSNR. Variant V3 disables the $\tau_{IFT}$ guidance, and while it achieves better LPIPS and PSNR compared to previous configurations, it lacks the adaptive temporal alignment that $\tau_{IFT}$ provides.

In V4, we evaluate the effect of removing the diffusion process entirely; while LPIPS remains low, PSNR drops, suggesting reduced pixel-level fidelity. V5 disables both diffusion and $\tau_{IFT}$, which results in a general performance degradation. Our full model (Ours), with all components enabled, achieves the best scores across both metrics, demonstrating the complementary nature of synthesizer, initial infill, temporal guidance, and diffusion-based refinement.

\begin{table}[t]
    \centering
    \resizebox{\textwidth}{!}{%
        \begin{tabular}{ccccc|cc}
            \toprule
            & Use Synthesizer & Initial Infill & Use of $\tau_{IFT}$ & Use Diffusion & LPIPS$\downarrow$ & PSNR$\uparrow$ \\
            \midrule
            V1 & No  & Yes & Yes & Yes & 0.0392 & 28.30 \\
            V2 & Yes & No  & Yes & Yes & 0.0367 & 28.81 \\
            V3 & Yes & Yes & No  & Yes & 0.0337 & 29.73 \\
            V4 & Yes & Yes & Yes & No  & 0.0341 & 29.48 \\
            V5 & Yes & Yes & No  & No  & 0.0345 & 29.54 \\
            Ours & Yes & Yes & Yes & Yes & \textbf{0.0327} & \textbf{29.79} \\
            \bottomrule
        \end{tabular}%
    }
    \caption{Results of the ablation experiment with 300 random samples, showing the performance of the proposed model variants on the ATD-12k-test dataset.}
    \label{tab:ablation}
\end{table}

\section{Conclusions}

We have described a novel diffusion-based video frame interpolation model with state-of-the-art performance on animated movies. One of its main features is that it estimates and encodes the temporal position of the intermediate frame, which is typically not well defined in traditional hand-made animation, because of the large variations resulting from the manual drawing.

The proposed architecture relies on the combination of standard flow-based, edge-aware warping methods with a deep learning model 

One line of research we want to pursue is to leverage the pixel-wise uncertainties over the reconstructed intermediate frames for developing a useful semi-automatic tool for animators to spot regions within the proposed interpolated images where some manual correction could be needed. 


\bibliographystyle{plain}
\bibliography{vfi}

\appendix

\section{Mathematical Details}
\label{apendix:A}

\subsection{Forward Marginal Distribution of $\mathbf I^{(t)}_{\tau}$}

As we define in \eqref{eq:forward distribution} the $\mathbf I^{(t)}_{\tau}$ can be sampled via the following equation:

\begin{equation}
\mathbf I_\tau^{(t)}=\mathbf I_\tau^{(t-1)}+\sum_{i=1}^n \alpha_i^{(t)} R_i(\mathbf I_\tau)+\kappa \sqrt{\sum_{i=1}^n \alpha_i^{(t)}} \epsilon^{(t)},
\end{equation}
where $\epsilon^{(t)} \sim \mathcal{N}(0, \boldsymbol{\mathrm{I}})$. By leveraging the sampling trick, we can establish a direct relationship between $\mathbf I_\tau^{(t)}$, which represents the progressively noisier version of our image at timestep $t$, and the original noiseless image $\mathbf I_\tau^{(0)} \triangleq \mathbf I_\tau$.

$$
\begin{aligned}
\mathbf I_{\tau}^{(t)} & =\mathbf I_{\tau}+\sum_{k=1}^t \sum_{i=1}^n \alpha_i^{(k)} R_i(\mathbf I_\tau)+\kappa \sum_{k=1}^t \epsilon^{(k)}\sqrt{\sum_{i=1}^n \alpha_i^{(k)}} \\
& = \mathbf I_{\tau} + \sum_{i=1}^n \eta_i^{(t)}R_i(\mathbf I_\tau) +\kappa \sum_{k=1}^t \epsilon^{(k)} \sqrt{\sum_{i=1}^n \alpha_i^{(k)}},
\end{aligned}
$$
where $\epsilon^{(k)} \sim \mathcal{N}(0, \boldsymbol{\mathrm{I}})$, so we can ``merge'' $\epsilon^{(1)}, \epsilon^{(2)}, \cdots, \epsilon^{(t)}$ into $\epsilon$, which variance is the sum of all the individual variance within the sum over $k$, and get

\begin{equation}
\mathbf I_{\tau}^{(t)} = \mathbf I_{\tau} + \sum_{i=1}^n \eta^{(t)}_iR_i(\mathbf I_\tau )+\kappa \sqrt{\sum_{i=1}^n \eta^{(t)}_i} {\epsilon}.
\end{equation}

Furthermore, the marginal distribution can be represented for $t=1,2, \cdots, T$:

\begin{equation}
q\left(\mathbf I_{\tau}^{(t)} \mid \mathbf I_{\tau}, \mathcal{J}\right)=\mathcal{N}\left(\mathbf I_{\tau}+\sum_{i=1}^n \eta^{(t)}_i R_i(\mathbf I_{\tau}), \kappa^2 \sum_{i=0}^n \eta^{(t)}_i \boldsymbol{I}\right).
\end{equation}

\subsection{Explicit form of $q\left(\mathbf I_{\tau}^{(t-1)} \mid \mathbf I_{\tau}^{(t)}, \mathbf I_{\tau}, \mathcal{J}\right)$}

We are going to start decomposing $q\left(\mathbf I_{\tau}^{(t-1)} \mid \mathbf I_{\tau}^{(t)}, \mathbf I_{\tau}, \mathcal{J}\right)$ using Bayes rule 

\begin{equation}
q\left(\mathbf I_{\tau}^{(t-1)} \mid \mathbf I_{\tau}^{(t)}, \mathbf I_{\tau}, \mathcal{J}\right) \propto q\left(\mathbf I_{\tau}^{(t)} \mid \mathbf I_{\tau}^{(t-1)}, \mathcal{J}\right) q\left(\mathbf I_{\tau}^{(t-1)} \mid \mathbf I_{\tau}, \mathcal{J}\right).
\end{equation}

We now focus on the form of the exponent of $q\left(\mathbf I_{\tau}^{(t-1)} \mid \mathbf I_{\tau}^{(t)}, \mathbf I_{\tau}, \mathcal{J}\right)$, namely,

$$
\begin{aligned}
\log q\left(\mathbf I_{\tau}^{(t-1)} \mid \mathbf I_{\tau}^{(t)}, \mathbf I_{\tau}, \mathcal{J}\right) = 
& -\frac{\left(\mathbf I_{\tau}^{(t)} - \mathbf I_{\tau}^{(t-1)} - \sum_{i=1}^n \alpha_i^{(t)} R_i(\mathbf I_{\tau})\right) \left(\mathbf I_{\tau}^{(t)} - \mathbf I_{\tau}^{(t-1)} - \sum_{i=1}^n \alpha_i^{(t)} R_i(\mathbf I_{\tau})\right)^T}{2 \kappa^2 \sum_{i=1}^n \alpha_i^{(t)}} \\
& - \frac{\left(\mathbf I_{\tau}^{(t-1)} - \mathbf I_{\tau} - \sum_{i=1}^n \eta^{(t-1)}_i R_i(\mathbf I_{\tau})\right) \left(\mathbf I_{\tau}^{(t-1)} - \mathbf I_{\tau} - \sum_{i=1}^n \eta^{(t-1)}_i R_i(\mathbf I_{\tau})\right)^T}{2 \kappa^2 \sum_{i=1}^n \eta^{(t-1)}_i} \\
= & -\frac{1}{2} \left[ \frac{1}{\kappa^2 \sum_{i=1}^n \alpha_i^{(t)}} + \frac{1}{\kappa^2 \sum_{i=1}^n \eta^{(t-1)}_i} \right] \mathbf I_\tau^{(t-1)} \mathbf I_\tau^{(t-1)T} \\
& + \left[ \frac{\mathbf I_\tau^{(t)} - \sum_{i=1}^n \alpha_i^{(t)} R_i(\mathbf I_\tau)}{\kappa^2 \sum_{i=1}^n \alpha_i^{(t)}} + \frac{\mathbf I_\tau + \sum_{i=1}^n \eta^{(t-1)}_i R_i(\mathbf I_\tau)}{\kappa^2 \sum_{i=1}^n \eta_i^{(t-1)}} \right] \mathbf I_\tau^{(t-1)T} + C \\
= & -\frac{\left(\mathbf I_\tau^{(t-1)} - \mu_t\right) \left(\mathbf I_\tau^{(t-1)} - \mu_t\right)^T}{2 \lambda^2} + C,
\end{aligned}
$$
where $C$ is a constant depending on terms that do \emph{not} contain any $I_\tau^{(t - 1)}$. Now we can identify the mean and standard deviation. Starting with the second one, we only have to take the coefficient corresponding to the quadratic term ($1/\lambda^2 = A$).

\begin{equation}
\lambda^2 = \sigma_t^2 =\kappa^2 \frac{(\sum_{i=1}^n\eta_i^{(t-1)})(\sum_{i=1}^n\alpha_i^{(t)})}{\sum_{i=1}^n\eta_i^{(t)}}.
\end{equation}

Now we use the deviation standard altogether with the linear term to identify the expression of the mean ($\mu = \lambda^2 B$),

\begin{equation}
\mu_t=\frac{\sum_{i=1}^n\eta^{(t-1)}_i}{\sum_{i=1}^n\eta^{(t)}_i} \mathbf I_\tau^{(t)}+\frac{\sum_{i=1}^n\alpha^{(t)}_i} {\sum_{i=1}^n\eta^{(t)}_i} \mathbf I_\tau - \Delta,
\end{equation}
where $\Delta$ is an expression that depends of the residuals $R_i(\mathbf \mathbf{I}_\tau)$'s, which can be seen as follows

\begin{equation}
\Delta = \frac{\left(\sum_{i=1}^n \alpha^{(t)}_i\right) \left(\sum_{i=1}^n \eta_i^{(t-1)} R_i(\mathbf{I}_\tau)\right)}{\sum_{i=1}^n \eta_i^{(t)}} - \frac{\left(\sum_{i=1}^n \eta_i^{(t-1)}\right) \left(\sum_{i=1}^n \alpha^{(t)}_i R_i(\mathbf{I}_\tau)\right)}{\sum_{i=1}^n \eta_i^{(t)}}.\label{eq:delta}
\end{equation}

Replacing \(\alpha_i^{(t)} = \eta_i^{(t)} - \eta_i^{(t-1)}\) and just working on the numerator leads to

$$
\begin{aligned}
\Delta \cdot \sum_{i=1}^n \eta_i^{(t)} &=\left[\left(\sum_{i=1}^n (\eta_i^{(t)} - \eta_i^{(t-1)})\right) \left(\sum_{i=1}^n \eta_i^{(t-1)} R_i\right) - \left(\sum_{i=1}^n \eta_i^{(t-1)}\right) \left(\sum_{i=1}^n (\eta_i^{(t)} - \eta_i^{(t-1)})R_i\right)\right]\\
&= \sum_{i=1}^n \eta_i^{(t)} \sum_{i=1}^n \eta_i^{(t-1)} R_i - \sum_{i=1}^n \eta_i^{(t-1)} \sum_{i=1}^n \eta_i^{(t)} R_i.
\end{aligned}
$$

Therefore, the reduced formula for $\Delta$ is:

\begin{equation}
\Delta = \sum_{i=1}^n \eta_i^{(t-1)} R_i(I_\tau) - \frac{\left(\sum_{i=1}^n \eta_i^{(t-1)}\right) \left(\sum_{i=1}^n \eta_i^{(t)} R_i(\mathbf I_\tau)\right)}{\sum_{i=1}^n \eta_i^{(t)}}.
\end{equation}

Hence, given that, we can get a common factor $\sum_{i=1}^n \eta_i^{(t-1)} / \sum_{i=1}^n \eta_i^{(t)}$ in the $\mu_t$ formula and we get the final result

\begin{equation}
\mu_t=\frac{\sum_{i=1}^n\eta_i^{(t-1)}}{\sum_{i=1}^n\eta_i^{(t)}} \left(\mathbf I_\tau^{(t)} 
 +  \sum_{i=1}^n \eta_i^{(t)} R_i(\mathbf I_\tau) \right)+\frac{\sum_{i=1}^n\alpha_i^{(t)}} {\sum_{i=1}^n\eta_i^{(t)}} \mathbf I_\tau - \sum_{i=1s}^n \eta_i^{(t-1)} R_i(\mathbf I_\tau).
\end{equation}

\end{document}